# Validation of the Scientific Literature via Chemputation Augmented by Large Language Models

Sebastian Pagel, Michael Jirasek, Leroy Cronin*

*School of Chemistry, The University of Glasgow, University Avenue, Glasgow G12 8QQ, UK.*

*Lee.Cronin@glasgow.ac.uk

**Abstract**

Chemputation is the process of programming chemical robots to do experiments using a universal symbolic language, but the literature can be error prone and hard to read due to ambiguities. Large Language Models (LLMs) have demonstrated remarkable capabilities in various domains, including natural language processing, robotic control, and more recently, chemistry. Despite significant advancements in standardizing the reporting and collection of synthetic chemistry data, the automatic reproduction of reported syntheses remains a labour-intensive task. In this work, we introduce an LLM-based chemical research agent workflow designed for the automatic validation of synthetic literature procedures. Our workflow can autonomously extract synthetic procedures and analytical data from extensive documents, translate these procedures into universal *X*DL code, simulate the execution of the procedure in a hardware-specific setup, and ultimately execute the procedure on an *X*DL-controlled robotic system for synthetic chemistry. This demonstrates the potential of LLM-based workflows for autonomous chemical synthesis with Chemputers. Due to the abstraction of *X*DL this approach is safe, secure, and scalable since hallucinations will not be chemputable and the *X*DL can be both verified and encrypted. Unlike previous efforts, which either addressed only a limited portion of the workflow, relied on inflexible hard-coded rules, or lacked validation in physical systems, our approach provides four realistic examples of syntheses directly executed from synthetic literature. We anticipate that our workflow will significantly enhance automation in robotically driven synthetic chemistry research, streamline data extraction, improve the reproducibility, scalability, and safety of synthetic and experimental chemistry.

**Introduction**

More than 65 million chemical reactions have been published in research papers and patents to date.[1] While being important cornerstone, databases like Reaxys, and the Open Reaction Database (ORD)[2] that attempt to capture this vast stream of data, are not sufficient to tackle the rapid validation and reproduction of reported data.[3] Moreover, almost 30 million new reactions have been added to the Reaxys database since 2014 alone, highlighting the need for improved ways of rapidly validating chemical reaction data beyond standardization of reporting practices and manual labour. Whereas 30 million data points might seem negligible in times when Video Generation Models and LLMs are trained on billions of examples.[4,5] This means the necessity of manual validation of chemical reactions creates an almost insurmountable challenge. Aside from the resource and time challenges associated with performing chemical reactions, finding missing parameters and reporting ambiguities, there are a vast number of esoteric experimental set ups and language variations reported. There is also an enormous backlog of unverified procedures with numerous new ones being reported every day which intensifies the problem. Commercial and Open-Source LLMs like GPT-4, or Llama have shown impressive abilities for textual comprehension and understanding ambiguous textual data.[5,6] Beyond pure text comprehension it was shown that these models exhibit excellent few-, one-, and zero-shot prediction abilities on unseen tasks without the need to finetune on specific tasks (in-context or gradient-free learning).[7,8] Furthermore, a plethora of *prompting techniques* has been developed, further improving the prediction abilities of these models.[9–12] This has led to vast amounts of research far beyond the field of Natural Language Processing into chemical research.[13–15] While machine-learning algorithms have been extensively contributed to advances in chemistry research,[16–20] the introduction of LLM as chemical agents has shown promising results to further automate chemical decision-making and experimentation and the *orchestrator* behind self-driving laboratories.[21–24]

Herein we present Autonomous Chemputer Reaction Agents (ACRA), a LLM based chemical multi-agent workflow, automating the tedious process of extracting reaction data from literature procedures, standardizing them, translating them into robotic instructions, conducting experiments and iterative *X*DL code development suggestions (Figure 1).

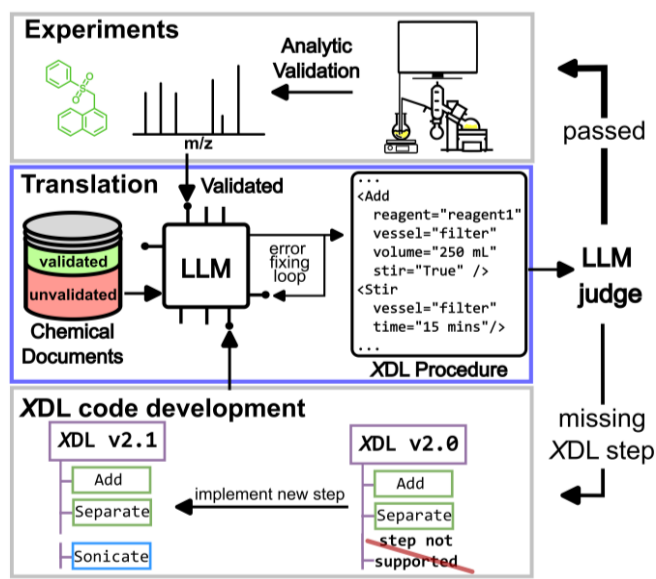

**Figure 1:** Conceptual overview of the workflow presented in this work for automated synthesis validation and iterative code development on demand of XDL.

ACRA is instructed to translate synthetic procedures into autonomously executable procedure via the Chemical Description Language (*X*DL), which represents synthetic steps, reagents, as well as available hardware in an unambiguous way (compare Figure 2C for a simplified *X*DL which is directly mapped to an abstract hardware configuration).[25] In previous work, it was shown how synthetic procedures represented in *X*DL can be used to unambiguously validate synthetic procedures[26]. While previous attempts were made to automatically translate literature procedures into *X*DL via semantic parsing[25] or similar to here, LLMs[22], none of these attempts enabled to go all the way from a literature document to the execution of a synthetic procedure, and importantly lack any way of validating that the generated *X*DL procedure matches the synthetic procedure. Starting from a literature source the *paper-scraping-agent* extracts synthesis procedures alongside purification information, analytical information and further related information by iteratively analysing parts of the text and creating a knowledge-graph (KG) of the given source. The *procedure-agent* subsequently sanitizes all extracted procedures, filling identified ambiguities with information from chemical databases and previously resolved ambiguities. Additionally, all procedures are classified in one of three categories (executable, reaction blueprint, or incomplete procedure). As executable and blueprint identified procedures are then translated into *X*DL by the *X*DL-*agent* and iteratively corrected with feedback from a three-stage validation workflow. Finally, successfully translated procedures are executed on a Chemputer platform. Synthetically validated procedures are than stored in a *X*DL database systematically increasing the number of validated and standardized procedures, greatly lowering the barrier of reproduction for other chemists. Additionally, the validated procedures and their associated *X*DL can be guide future translation. All generated data and extracted data are stored in a unified *labbook* (see Figure 2A/B and SI section 2 for details). The *long-term memory storage* allow ACRA to *learn* from previous experiments by providing previously translated examples as well as resolved ambiguities within the context of the prompt. Though ACRA was mostly tested for handling English literature procedures, it showed impressive cross-language capabilities, when parsing foreign language documents and procedures, potentially lowering language barriers in scientific reporting. To showcase the potential of ACRA, we demonstrate the automatic, parsing, translation, and sanitization capabilities on four reactions. This way, we were able to show that ACRA can autonomously parse literature for relevant and executable procedures, plan, translate them into *X*DL and finally execute the translated procedures, showing robustness to language ambiguities, different languages, and even potential reporting errors.

**Results**

**Extracting synthesis information from chemistry literature**

The vast amounts of data contained in scientific publications, regarding compound properties, reactivity, reaction execution and product analysis can be spread over 10s to 100s of pages typically divided into main publication article and supporting information. While the main publication usually contains a higher-level description of the performed experiments and most relevant results, spectroscopic analysis data, like nuclear magnetic resonance- or mass-spectrometry, is usually hidden in the supporting information alongside potentially relevant additional information to accurately recreate the experiments.

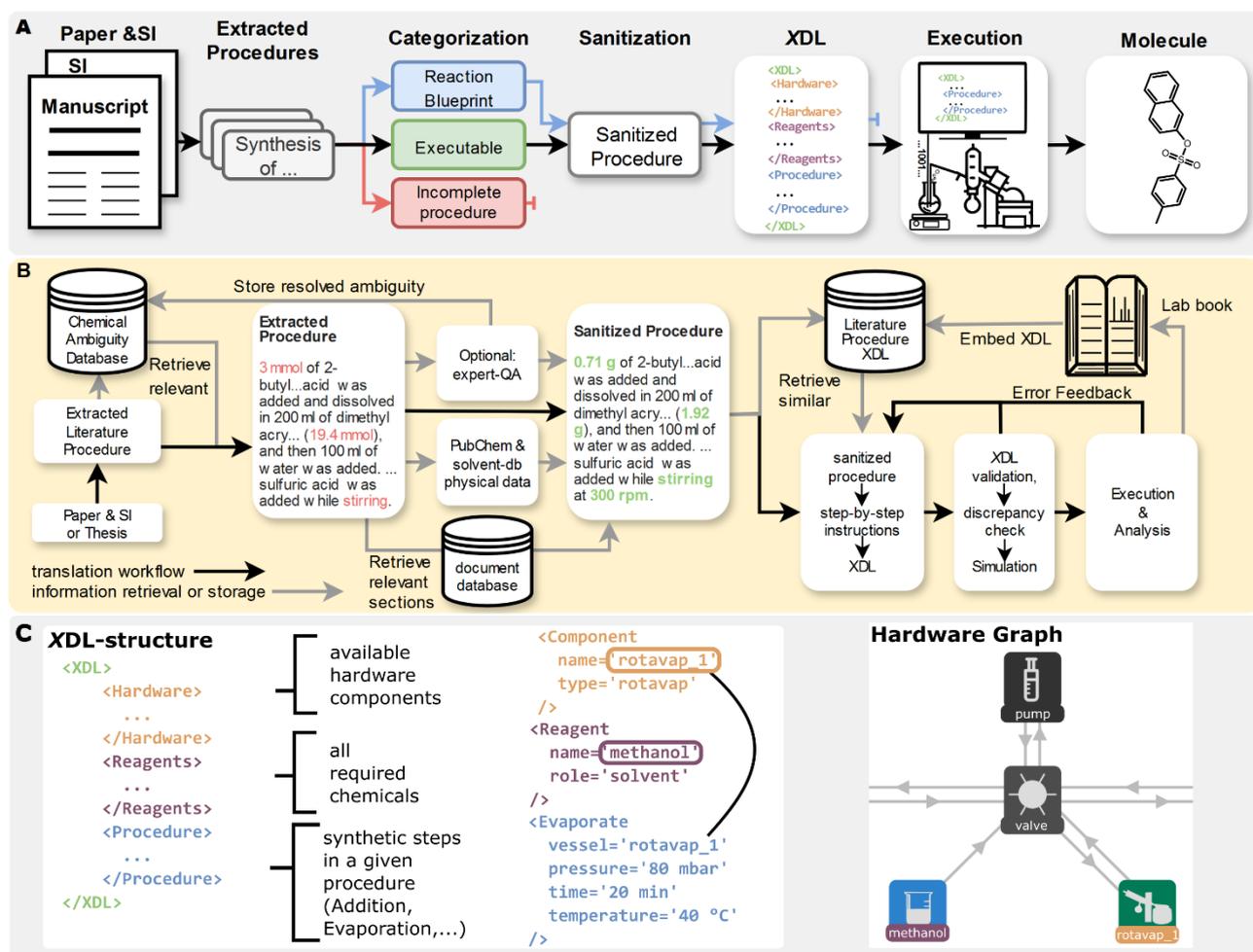

**Figure 2: Overview of the proposed framework for automated extraction, translation, and validation of synthesis procedures** **A)** Simplified overview of the proposed framework for extraction, sanitization, translation and validation of chemical reaction procedures using LLM-based agents. **B)** Detailed depiction of the flow of chemical procedures from literature to robotic execution. First, entire papers (and their supporting information/ or any other text document) are parsed by a *scraping-agent*, and all synthesis-related data is extracted to a knowledge graph. All extracted procedures alongside relevant chemical ambiguities and physiochemical data are passed to a second agent (*procedure-agent*) to sanitize the procedure (fill in missing physiochemical information etc.). The sanitized procedure is categorized by the *procedure-agent*, and subsequently translated by the *X*DL-*agent* into *X*DL. The translated procedure is (if needed) iteratively improved by a three-step sanitization pipeline. Finally, the validated *X*DL is stored after optional physical execution and analysis, alongside the extracted data, into a *labbook*. The validated *X*DL is embedded and stored in a vector database, which is used as a long-term memory of the *X*DL-*agent* for future translations. **C)** Simplified description of a XDL procedure, and depiction of the hardware graph to execute the given procedure.

To harvest this data, the first stage of ACRA (*scraping-agent*) parses a given literature text (and its supporting information or any other documents if provided) and extracts data into a knowledge graph (KG) of relevant synthesis-related information to execute and analyse the described procedures (Figure 3A and D). ACRA parses literature resources by first chunking a given text into 4096 token fragments and then iteratively extracts and combines the data until all text has been parsed (SI section 1). The *scraping-agent* is instructed to extract all chemical names with their abbreviations and synonyms, procedure texts, purification data, as well as analytical data and additional information, anywhere in the documents. Later, during translation of the extracted procedures in *X*DL an additional translation graph is created which can be linked to the KG via the procedure title (see below and Figure 3D). Additionally, the initial text document is embedded into a vector-database in chunks of 2048 tokens and referenced in the knowledge graph for later retrieval during translation of the procedure, if additional information is required. This way, ACRA can autonomously extract procedure descriptions, chemical information, and analytical data as well as any pitfalls or limitations highlighted by the authors from vast amounts of text. To test the extraction of synthesis-related information and the construction of the KG we executed the literature extraction and KG generation module on 10 scientific publications and an organic chemistry PhD thesis (SI section 2). To test the cross-language capabilities, we also tested it on an organic chemistry undergraduate practical transcript written in German. In total 717, 57, and 117 procedures were extracted from the scientific publications, a PhD thesis, and German practical script, respectively. To estimate how many of those procedures contain all the required information to reproduce a given procedure, ACRAs *procedure-agent* was executed aiming to resolve any ambiguities and fill in missing data for the unambiguous translation to *X*DL (SI section 1). During the sanitization, the *procedure-agent* categorizes the extracted procedure into 'executable', 'blueprint' (general procedure descriptions etc.), and 'incomplete'. Out of the 717 extracted procedures from the scientific publications, 427 were marked as executable, 89 as blueprint, and 201 as containing missing information. The procedures from the German undergraduate practical and PhD thesis categorization were split in 48, 2, 7 and 93, 6, 18, respectively (Figure 3C).

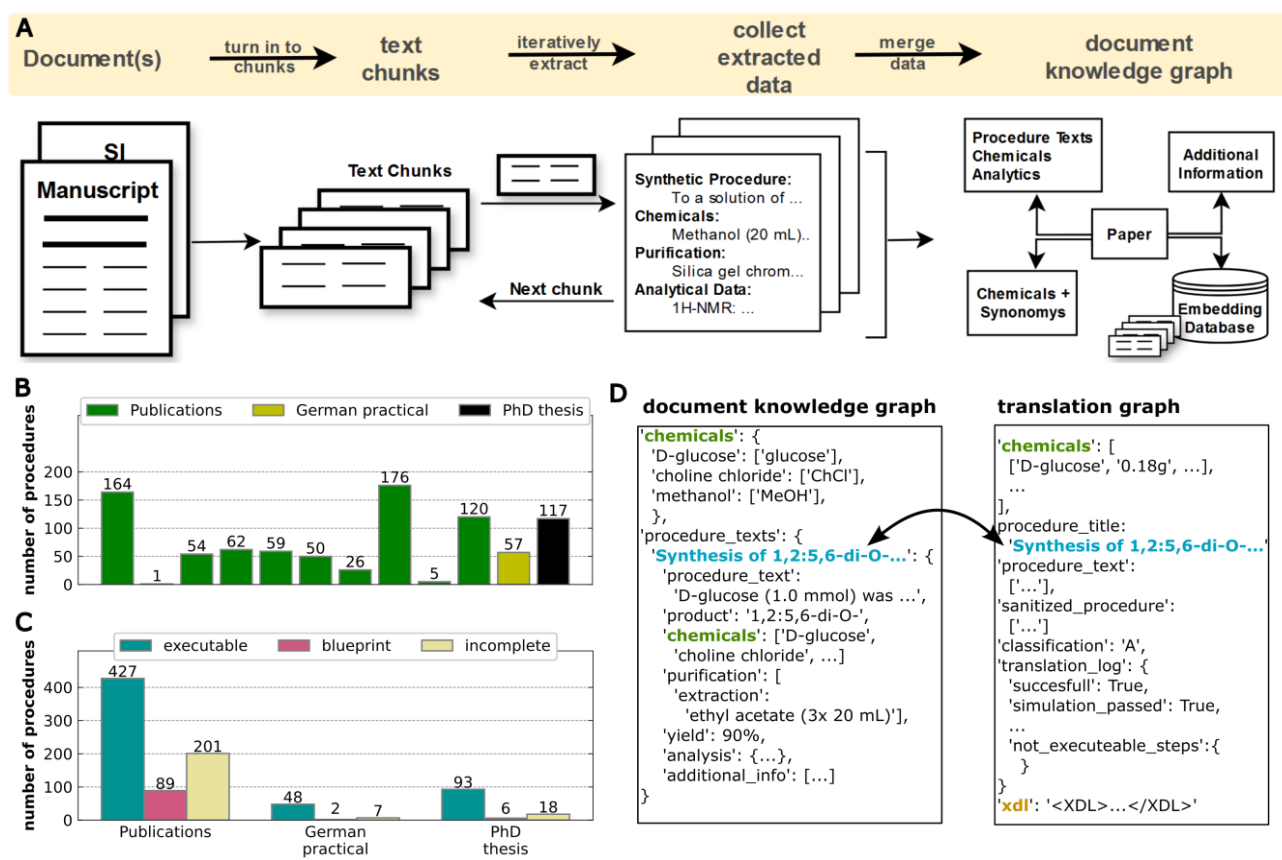

**Figure 3: Synthesis data extraction from chemical literature documents. A)** Synthesis data is extracted from documents containing chemical synthesis information and turned into a knowledge graph, by first extracting all textual data, chunking the text, iteratively extracting synthesis-related data in a JSON-format, and finally combining all extracted data into a combined data structure (knowledge graph). The *scraping-agent* is instructed to extract all chemical names with their abbreviations and synonyms, procedure texts, and purification data, as well as analytical data. Additionally, the initial text-document is stored and embedded into a vector-database, and referenced in the knowledge graph, for later retrieval during translation of the procedure, if additional information is required. **B)** Benchmark of extracting synthesis procedures from different document types. 10 publications (compare SI 2.1), a script from an undergraduate organic chemistry practical in German language, as well as an organic chemistry PhD thesis, were used to test the extraction capabilities. In total 729 procedures were extracted from the publications, 57 from the organic chemistry practical transcript, and 117 from the PhD thesis. **C)** Categorization of the extracted procedures by the procedure agent into executable, blueprints (general procedures etc.) and incomplete procedures. **D)** Simplified depiction of a document knowledge graph linking retrieved information to a translation graph, created during procedure to XDL translation (see below).

**Precise and executable *X*DL procedures via validity check, discrepancy analysis, and hardware-constrained simulation**

While extraction of relevant data from largely unstructured literature procedures and sanitisation of procedures are a significant step in automating chemical synthesis validation, accurate translation of literature procedures into unambiguously executable robotic instructions remains a challenge. The Chemical Descriptor Language *X*DL has been used to unambiguously, and reproducibly execute and share chemical reactions in a hardware-agnostic manner[25–29]. While translation from literature procedures to *X*DL has been presented before,[22,25] there remains a significant bottleneck in the accurate translation and automatic validation of these translated procedures. Whereas syntax errors have been used before to iteratively improve upon previously generated *X*DL instructions,[22] we found that a substantial proportion of translated procedures remained syntactically erroneous, had missing steps and were not sufficiently validated on a realistic robotic setup. To improve upon these shortcomings, we implemented a validation pipeline that first creates valid *X*DL with error feedback from a *X*DL-parser that can simultaneously find all syntactic errors in a given *X*DL. The *X*DL is then scrutinised by a *critique-agent* instructed to find any discrepancies in steps in the *X*DL that were mentioned in the literature procedure and implement them subsequently (typically referred to *LLM-as-a-judge*[30]). Finally, the generated *X*DL is mapped to a predefined robotic platform (Chemputer, SI section 6) and the execution of these steps is executed in simulation, constrained by the robotic platform (Figure 4A). The *X*DL parser captures syntactical issues, ill-defined physical units (e.g. temperature, and pressure units), and missing hardware or chemical reagents. Simulation of the physical execution was tested via computational simulation of the produced procedure, capturing errors such as invalid temperature or rotation speed ranges. While these checks help ensure the executability of a translated procedure, they do not ensure the completeness or accuracy of the translated procedure. Importantly, the *critique-agent* proves to be a vital part of the accurate translation of syntactic procedures, identifying missing or miss ordered steps. Examples for each of the three-stage feedback responses are shown in Figure 4B. To translate literature procedures into *X*DL, the *X*DL-agent (SI section 1.1) was instructed to first, extract all chemicals and their role in the procedure (e.g. solvent or catalyst), then decompose the procedure into step-by-step instructions and translate them into *X*DL in a *ReAct-style* response format [31], and finally combine the individual steps into a single *X*DL, within a single prompt (Figure 4C). During the iterative improvement, the *X*DL-agent is instructed to first, map the errors identified from the validation stages to the part of the *X*DL procedure causing the error, and finally correct the corresponding lines. In each iteration of the translation the five most similar synthetic procedure-*X*DL pairs are provided within the prompt allowing to *X*DL-agent to learn from previously translated and validated procedures (see below and SI section 1.2). Additionally, the *X*DL documentation, and previously resolved ambiguities are provided within the context of the prompt (SI section 1.1 and 1.2 and below).

To test the performance of the validation stages, 150 procedures (three times 50 independently sampled), were translated into *X*DL procedures (SI section 2.4). The maximum number of iterations (*X*DL-generation → Validation → Feedback → *X*DL-generation) for the generation of error-free *X*DL after all three validation stages was set to 6 (see Figure 4E for distribution of actual number of iterations). 99.33% of procedures were translated into valid *X*DL (passing the *X*DL-validity check) and 94.67% of procedures were additionally successfully validated during procedure-*X*DL discrepancy check and simulation of the execution, ensuring their accuracy and executability of a suitable platform (Figure 4D). This highlights the necessity for validation beyond the mostly syntactical validation presented in previous studies to generate accurate and executable *X*DL procedures.

**Using long-term memory storages and past experiments to learn from experience**

Though reported synthetic procedures usually contain sufficient information for an expert chemist to infer implicitly assumed information, autonomously identifying missing or assumed information can pose a significant problem in autonomous synthesis execution. To tackle this problem, we equipped

ACRA with an ambiguity database containing implicit knowledge expert chemists identified in previously published literature procedures. To initialize the database 5 synthetic procedures were carefully annotated and described and each piece of the procedure with its detailed explanation was stored in a vector database (Chemical Ambiguity Database).

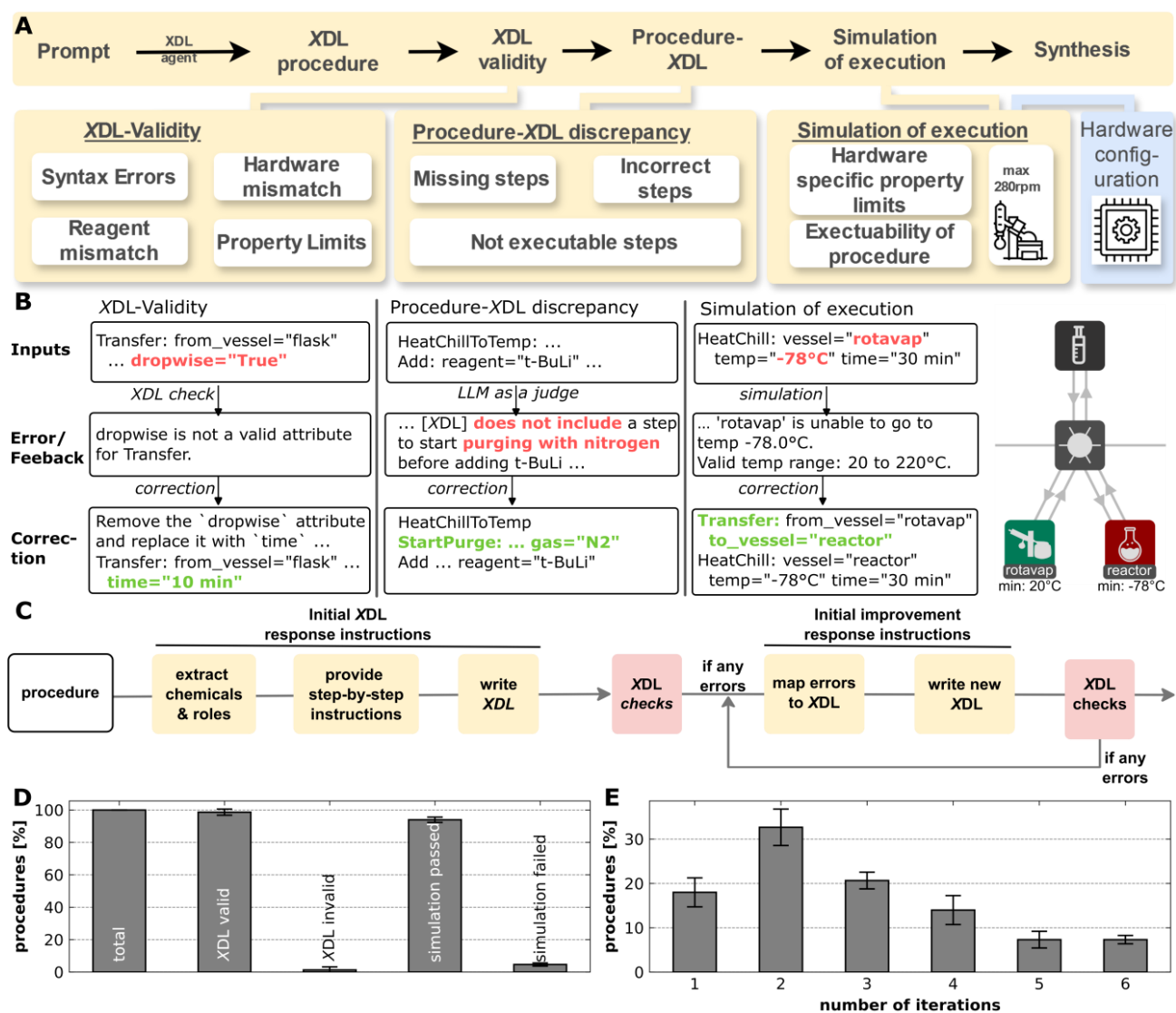

**Figure 4: Generating accurate and executable *XDL*-procedures for automatic synthesis execution. A)** Overview of workflow from synthesis procedure to validated *XDL* and synthesis. The procedure is translated into *XDL*, checked for errors, analysed to find discrepancies between the natural language procedure and the *XDL*, and finally simulated in a hardware-constrained environment. All errors captured along this pipeline are iteratively fed back to the LLM-agent to correct the errors. **B)** Examples for the three stages in which the *XDL* is scrutinized. **C)** Overview of response instructions for the *XDL* agent within ACRA and iterative improvement of the *XDL*s using the errors from the three stages described in **A). D)** Statistics of a total of 150 randomly selected procedures (three times 50 independently sampled), on passing the stages described in **A). E)** Number of iterations to completion of the translated procedures using the above-described workflow.

The pieces of the procedures that were explained were embedded into a 2048-dimensional vector to enable semantic search in case pieces of a new procedure need further clarification with the embedding-model *text-embedding-large* from OpenAI. During the sanitization of procedures relevant parts from this Chemical Ambiguity Database (CAD) serve as a long-term memory containing chemical intuition or implicit information that is usually not explicitly captured in textual form and learned by synthetic chemists during their educational programs. Relevant information contained in the CAD is retrieved by embedding sentences of a procedure into the same 2048-dimensional vector space and the semantically most similar information is selected as estimated by the cosine-similarity of their embedding-vectors. This information is then included within the context of a prompt (SI section 1.1). Additionally, ACRA can optionally ask questions about any parts of a given procedure to an expert chemist, whose answer, alongside the section of the procedure in question is stored in the long-term ambiguity database available for subsequent executions (SI section 1.1 and 1.2). This information together with molecular information extracted from PubChem about the chemicals used in each procedure (i.e. conversion of chemical names to IUPAC, and g/mmol to g/mol) and a local solvent database (containing boiling points etc.), is then provided to the *procedure-agent* to prepare a procedure with all necessary information to be translated into *X*DL. During the translation of sanitized procedures into *X*DL, the 5 most similar previously translated synthetic procedures and the corresponding *X*DLs are provided within the context of the prompt to help translation, usually referred to as *few-shot prompting* via *Retrieval Augmented Generation* (RAG; SI section 1.2).

To evaluate the influence of the *memory* and data components, 75 procedures were translated into *X*DL with different parts of the workflow removed. First, to fully test ACRAs capability to translate synthetic procedures to *X*DL, we initialized the *X*DL database with 62 procedures of previously published synthetic procedures-*X*DL pairs[28]. Each successfully translated procedure-*X*DL pair will be added to this database and is thus available for subsequent translations. We found that ACRA successfully translated 100% of the procedures into valid *X*DL (*X*DL validity check), and 6.7% failed to pass at the later stages (discrepancy check and simulation of execution). To test the influence of removing the *XDL* database, the procedures were additionally translated into *XDL* once, without any initial *XDL*s in the database but with continuous addition of successful translations, and once completely without the *XDL* database. While in both cases 100% of the procedures could be translated into valid *XDL*, 9.3% and 10.7% of procedures did not pass the latter two stages of the validation pipeline, respectively (Figure 5B). In a last test, all databases, and external data sources were removed from the translation process.

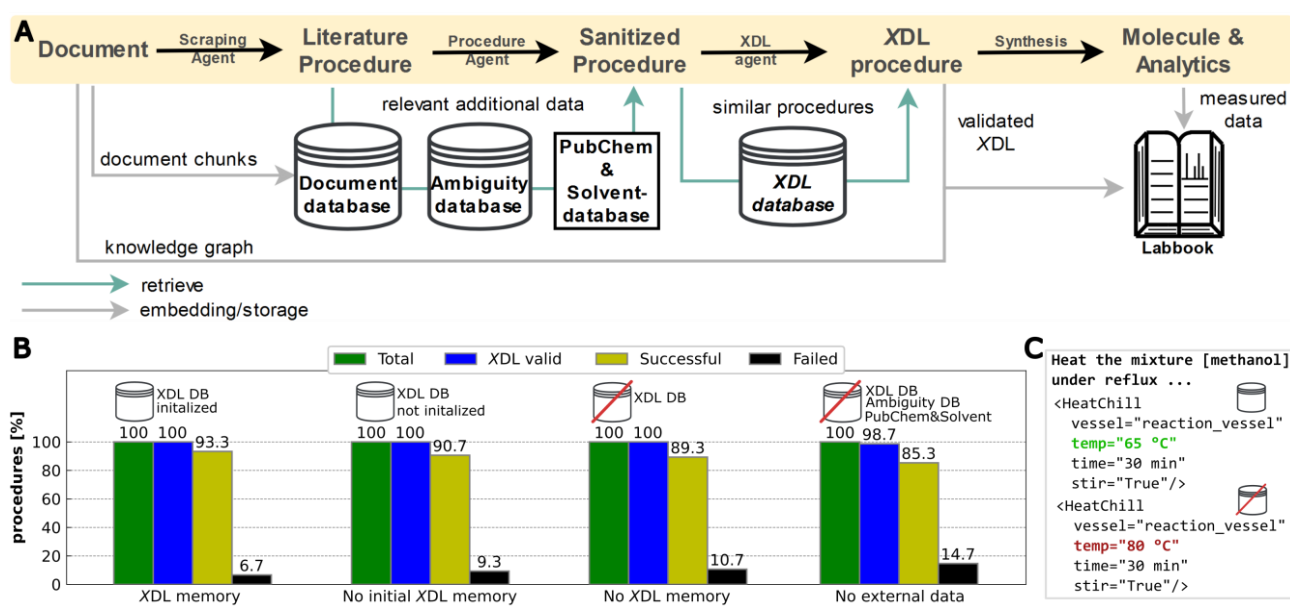

**Figure 5: Overview of inclusion of external data sources and long-term memory components in the translation of synthetic procedures to XDL. A)** Given a document containing the description of a synthetic procedure, the document is embedded in chunks of 2048 tokens into a vector database. During translation of extracted procedures, ACRA is instructed to ask questions that may be present in the document, but not the procedure (e.g. general procedure instructions) which are then retrieved from the vector database and included in the prompt. Additionally, previously resolved ambiguities are included, as well as physiochemical information about the identified chemicals in the procedure. During the XDL generation the 5 most similar, previously validated XDLs are retrieved from a vector database and included within the prompt. The combined data from the original document, a reference to the document database, resolved ambiguities, the validated XDL and any analytical data from the original procedure are finally stored within a virtual *labbook*. **B)** Influence of inclusion (or exclusion) of different storage and external data on the translation success of literature procedures to XDL. While in all stages, valid XDL was generated in almost 100 percent of cases, the overall rate of success increased from 85.3% to 93.3% by the inclusion of the different data and memory sources. **C)** Translation example where exclusion of data sources leads to a validated *XDL* but set too high temperature for refluxing of a methanol-based solution.

This way, no *X*DL database, no Chemical Ambiguity Database, and no additional chemical information was provided for the translation. The document database will only be used, if translation starts from a document, and was thus not used by default in this experiment. While 98.7% of all procedures could still be translated into valid *X*DL, 14.7% of procedures failed in one of the latter stages (discrepancy check or simulation of execution). This shows, how the use of previously translated and validated examples can be used to systematically improve the translation capabilities of LLM-based *agents* in chemistry without the need for any further training of the underlying models. The use of additional external sources like a curated chemical database proves to be a valuable addition to increasing the success rate of translated procedures. Additionally, these helps prevent hallucination when physically accurate values are required (e.g. refluxing temperatures; Figure 5C).

**Systematically improving *X*DL by identifying not executable synthetic steps**

Though a wide variety of reactions have been demonstrated to be executable using *X*DL procedures, *X*DL is still actively developed to add support of an increasing number of synthetic operations. Additionally, while most newly published procedures will not contain logically new steps, synthetic chemistry itself is changing as well requiring new synthetic steps. To identify and suggest new steps that should be added to the *X*DL standard, the *critique-agent* described above, was additionally to identify missing steps and instructed to identify steps that are currently not executable within *X*DL during translation of synthetic procedures. Steps that were identified as not executable were collected, clustered, and analysed to provide suggestions for new *X*DL steps (Figure 6A and B). Suggested *X*DL steps can then be analysed and finally integrated based on urgency and ease of integration for the next version of *X*DL. To provide a roadmap to systematically improve *X*DL and make it universally applicable, 350 as not executable marked steps were analysed which were collected throughout this work. The steps were clustered resulting in 26 new feature suggestions for future generations of *X*DL (Figure 6C) showcasing how this workflow can be used to systematically identify relevant new step suggestions. Additionally, 65 million procedures from the Reaxys database were analysed on the synthetic keywords identified and provided within the database.

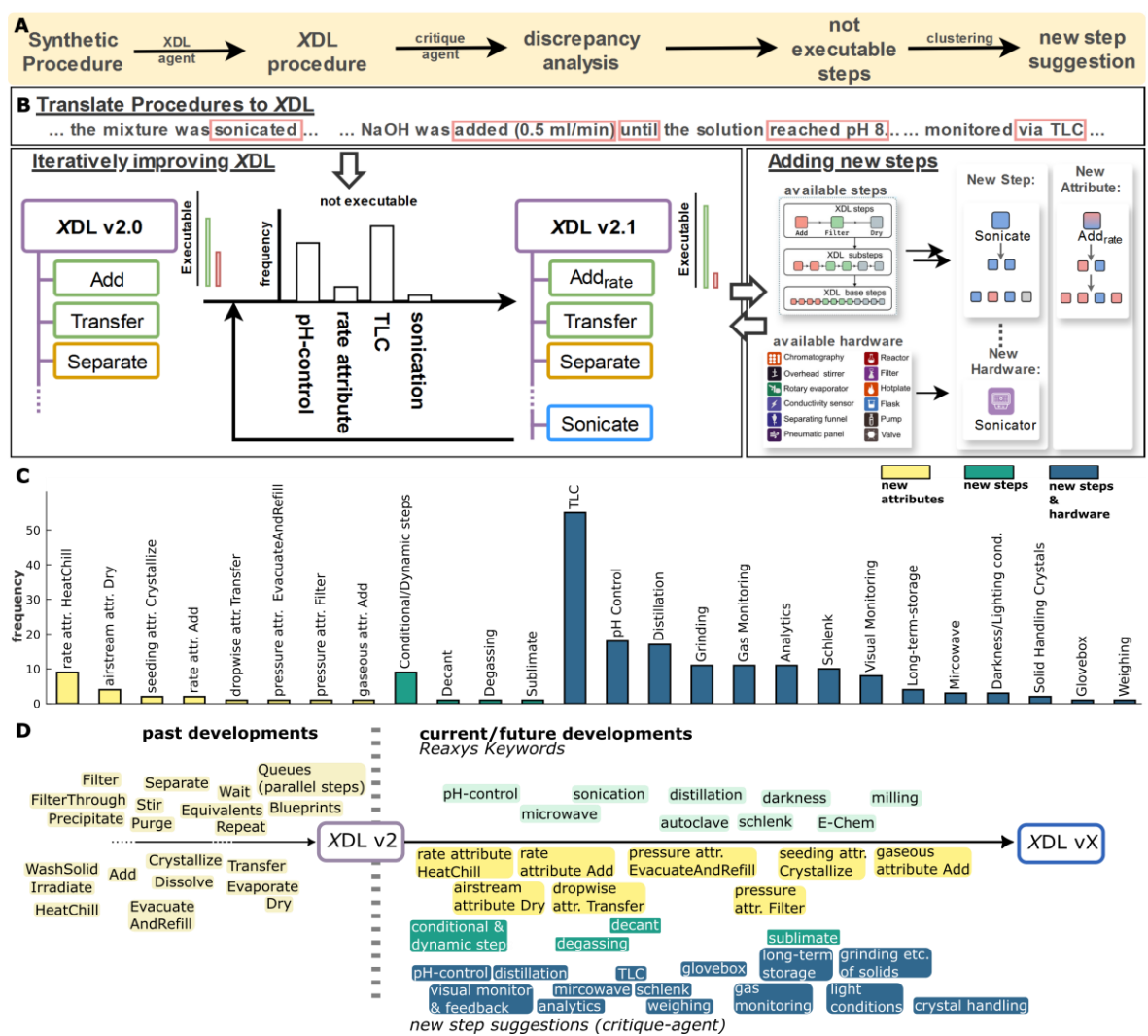

**Figure 6: Capturing currently unsupported steps from synthetic procedure and creating a roadmap to systematically more universal *X*DL. A)** Developed workflow to capture steps that are not executable in the current version of *X*DL. Synthetic procedures get translated into *X*DL as described above. The critique-agent then checks the translated procedure and the original procedure and captures steps that can currently not be translated into *X*DL steps. The combined not executable steps from a set of translated procedures are than clustered and used to plan new steps for future version of *X*DL. **B)** Conceptual overview of how not executable steps are used to iteratively improve *X*DL. Not executable steps from a batch of procedures are first clustered, categorized, and subsequently depending on urgency and ease of implementation included in the *X*DL language. Depending on the update this results in new attribute or require new step and hardware integration. **C)** Classification of not executable steps in this work into potential new *X*DL steps or features grouped by type of modification required for implementation (new attribute, steps, or step and hardware support). **D)** Overview of *X*DL development and supported steps. The 26 newly suggested steps are grouped by type and sorted urgency and ease of implementation from left to right. Additionally, to the not executable steps identified as specified above, ~65 million procedures from the Reaxys database were analysed on their most frequent keywords, and the top 1000 keywords clustered to give new step suggestions.

The 200 most frequently identified keywords were analysed and grouped into 20 categories (see SI section 4.3). Categories for which no abstract *X*DL step exists in the current version of *X*DL are shown in Figure 6D. Comparison of the newly suggested steps as an outcome of the not executable steps from the *critique-agent* and those from keywords from Reaxys procedures shows that the not executable steps result in substantially more specific suggestions ranging from new attributes (e.g. rate-control for heating or cooling steps) to steps requiring new hardware implementations (e.g. automated TLC analysis). The 26 new *X*DL feature suggestions grouped by type of adjustment required for implementation and ranked by ease of implementation and urgency resulting in a suggested road-map for future *X*DL generations (Figure 6D).

**Paper to Molecule**

To showcase the capabilities of ACRA in helping to automate chemical synthetic literature validation all the way from parsing a literature source to executing a synthetic procedure, we tested it on two English synthetic procedures, one German synthetic procedure, as well as one scientific publication, covering a variety of document types and use cases. The two English procedures detailed the synthesis of p-toluenesulfonate[32] and 2-Methyl-2-(3-oxopentyl)-1,3-cyclohexanedione[33] (Figure 7 B and C) and were translated into 23 and 20-step *X*DL procedures respectively. In the first, refluxing temperatures were correctly assigned, and volumes for adjustment of pH estimated since no adjustment of pH step was implemented in *X*DL version for this study. Though the estimation of the addition of volumes for adjustment of pH might be quite sensitive to errors, ACRA notably labelled the step adjustment of the pH as a not exactly executable step, as described above. The German synthetic procedure described the synthesis of 3-Methoxy-3-oxopropanoic acid. During the translation, all chemical names, as well as the synthetic procedure, were correctly translated. Additionally, the procedure specified lifting the reaction vessel halfway from the oil bath which is not a directly executable step in *X*DL. ACRA reasoned that the reaction vessel, cannot be lifted [from the heating vessel] and instead reduced the temperature from 65°C to 50°C, accurately interpreting the intention of the step. To showcase the potential of automatic execution of literature procedure ACRA was lastly provided a scientific publication detailing the synthesis of multiple sugar compounds. 5 synthetic procedures were extracted from the procedure. Two of the procedures were classified as blueprints (they were general procedures) while the other three were correctly classified as executable. One of the procedures was selected for validation describing the synthesis of Methyl 4,6-O-benzylidene-α-D-glucopyranoside[34]. Notably, the procedure mentioned extracting 30 times with 20 ml ethyl acetate, appearing to be reporting an error in the procedure. During the translation, ACRA identified this to be changed to 3 instead given the context of the paper and previously translated examples without further instructions. The translated procedure was executed, but no conversion of starting materials could be detected. The procedure was additionally executed with varying hardware implementations by a synthetic chemist closely following the reported procedure. Nevertheless, the reported reaction could not be verified without substantial alterations of the procedure and was thus classified as not reproducible.

**Conclusion**

In this work, we demonstrated how LLM can be used to autonomously validate chemical synthesis literature from parsing of a literature text to final execution of the synthesis. Importantly, we showed how syntactic validation of generated protocols for the robotic execution of synthesis procedures in the *X*DL is not sufficient for the accurate translation of literature procedures. Whilst simulating a chemical reaction at an atomistic scale remains a challenge to be conquered, we demonstrate, how simulation of the robotic execution in a hardware-constrained manner can be used to generate hardware-specific *X*DL procedures that are directly executable. We hypothesize that this concept can be further developed to integrate more simulation data in LLM-based validation workflows and lead to even improved efficiency and accuracy, ultimately helping to run entire laboratories with LLM-based agentic systems as their *brains*.

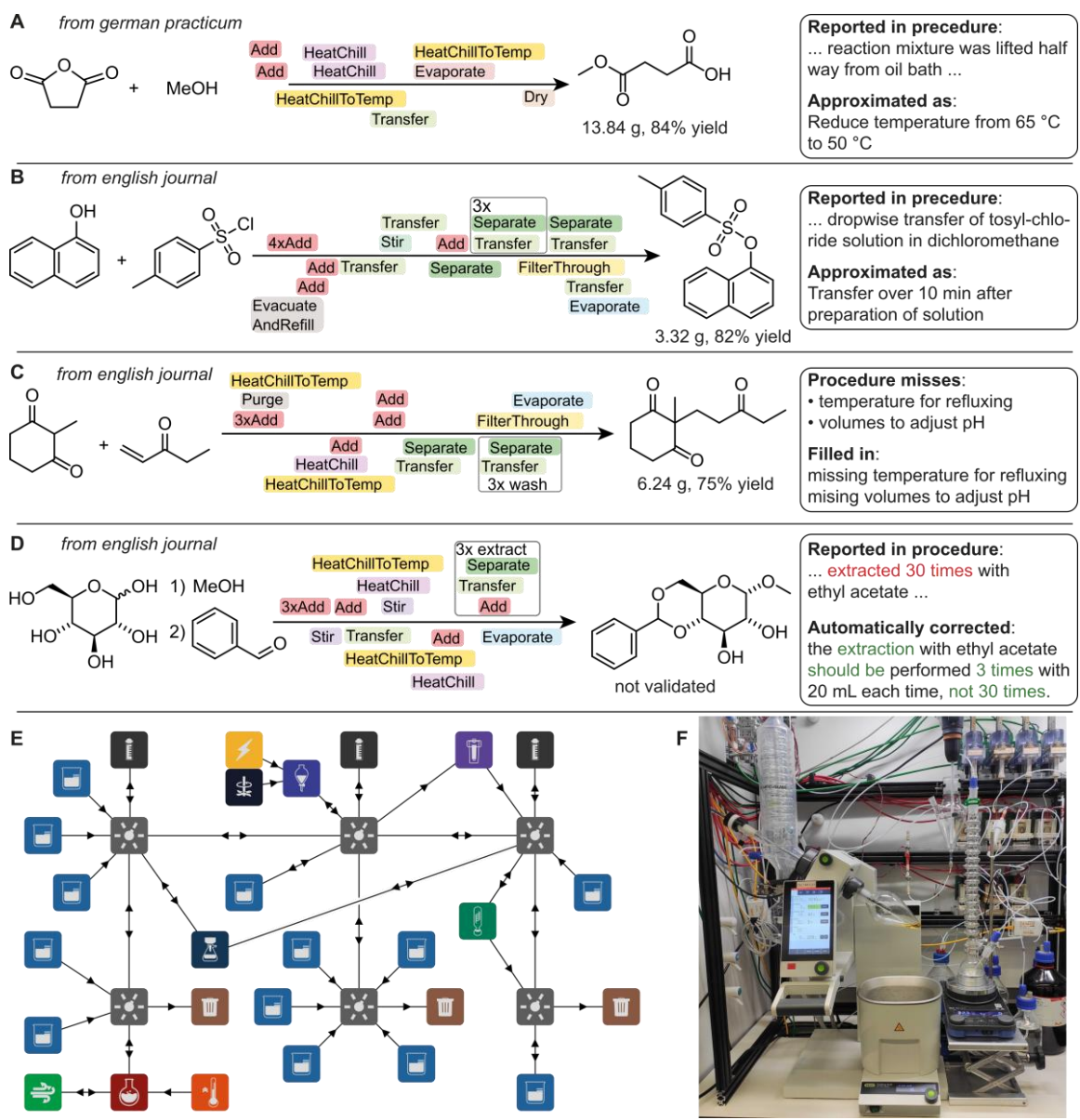

**Figure 7: Synthesized molecules from *X*DL procedures generated by ACRA on a Chemputer platform. A)** Synthesis of 3-Methoxy-3-oxopropanoic acid. A procedure in German for the synthesis of 3-Methoxy-3-oxopropanoic acid was provided to ACRA, automatically translated and adapted to be executable on a Chemputer platform (SI section 5.1). **B)** Synthesis of p-toluenesulfonate via a 23-step *X*DL procedure generated from a literature procedure (SI section 5.2). Notably, the dropwise transfer of solutions was approximated as a transfer of 10 minutes. **C)** Synthesis of 2-Methyl-2-(3-oxopentyl)-1,3-cyclohexanedione from a literature procedure (SI section 5.3). Temperatures for refluxing were correctly assigned, and volumes to adjust the pH were estimated, pH-control was not available on the given physical platform. **D)** Synthesis procedure of methyl 4,6-O-benzylidene-α-D-glucopyranoside directly parsed from a research article detailing the synthesis procedure via a 22-step *X*DL procedure. Notably, the procedure mentioned extracting the reaction mixture unusually many times (30 times), which ACRA automatically changed to 3 times, given the context of the paper (SI section 5.4). The procedure was executed 3 times with various cautions hardware implementations and was concluded to be not reproducible without significant alteration of the procedure / including more details in original procedure. For more details confer SI section 5.4. **E)** Example of Chemputer graph connectivity for one of the experimentally validated syntheses. **F)** Photo of Chemputer used for experimental *X*DL validation.

Similarly to how a human chemist might learn from previous examples, we showcase how integrating previous experiences, and translations into the translation workflow helps to substantially improve the translation success rate. Expanding on this concept might thus be an important step into open-ended chemical discovery. This requires an unambiguous and universal way of representing chemical procedures (compare Figure 7D). To improve on existing paradigms, we showcase how feedback from an LLM can be used to systematically suggest missing operations (here *X*DL steps) to fully cover all operations required to perform chemical experiments. We identified 26 new *X*DL feature suggestions and created a potential roadmap for future implementation.

**Methods**

The *critique-, X*DL-, and *procedure-agents* used GPT4o for all prompts. Only the *scraping-agent* used GPT4o-mini for the reduced cost, and increased token-output size. All *embeddings* were generated with the OpenAI model *text-embedding-large*. All code was written in Python3 with standard libraries apart from the *X*DL and Chemputer-specific libraries. The default *X*DL library (https://gitlab.com/croningroup/chemputer/xdl) was modified to capture all syntactic errors in parallel. The Chemputer*X*DL was modified to automatically map generated *X*DLs to a predefined hardware graph. *X*DL files (.xdl) and Chemputer graph files (.json) can be viewed and edited with the ChemIDE app on https://croningroup.gitlab.io/chemputer/xdlapp/. The *X*DL software standard is linked here: https://croningroup.gitlab.io/chemputer/xdl/standard/index.html. All *X*DL and Chemputer specific software packages can be made available upon reasonable request.

**Data Availability**

The code written for the implementation of ACRA will be available after publication in a peer reviewed journal at https://github.com/croningp/acra.

**Acknowledgements**

We acknowledge financial support from the John Templeton Foundation (grant nos. 61184 and 62231), the Engineering and Physical Sciences Research Council (EPSRC) (grant nos. EP/L023652/1, EP/R01308X/1, EP/S019472/1 and EP/P00153X/1), the Breakthrough Prize Foundation and NASA (Agnostic Biosignatures award no. 80NSSC18K1140), MINECO (project CTQ2017-87392-P) and the European Research Council (ERC) (project 670467 SMART-POM). We like to acknowledge Dean Thomas for the feedback on the manuscript and many helpful discussions.

**Author contributions**

L.C. conceived the idea and research plan together with S.P. and M.J. S.P. built and developed the workflow for the implementation of ACRA module with contribution of M.J. S.P. and M.J. implemented the experimental setup. M.J. and L.C. mentored S.P. S.P. wrote the manuscript with contributions from all authors.